\documentclass[sigconf]{acmart}

\AtBeginDocument{%
  \providecommand\BibTeX{{%
    \normalfont B\kern-0.5em{\scshape i\kern-0.25em b}\kern-0.8em\TeX}}}

\usepackage{threeparttable}
\usepackage{xspace}
\usepackage{multirow}
\usepackage{multicol}
\makeatletter
\DeclareRobustCommand\onedot{\futurelet\@let@token\@onedot}
\def\@onedot{\ifx\@let@token.\else.\null\fi\xspace}

\makeatletter
\DeclareRobustCommand\onedot{\futurelet\@let@token\@onedot}
\def\@onedot{\ifx\@let@token.\else.\null\fi\xspace}

\def\ie{\emph{i.e}\onedot} 
 
\def\etc{\emph{etc}\onedot}

\makeatother

\begin{document}

\title{psPRF:Pansharpening Planar Neural Radiance Field for Generalized 3D Reconstruction Satellite Imagery}


\author{Tongtong Zhang}
\email{tongtong_zhang@sjtu.edu.cn}
\orcid{0000-0002-4797-4860}
\affiliation{%
  \institution{School of Aeronautics and Aestronautics, Shanghai Jiao Tong University}
  \city{Shanghai}
  \country{China}
  \postcode{200240}
}
\author{Yuanxiang Li*}
\email{yuanxli@sjtu.edu.cn}
\affiliation{%
  \institution{School of Aeronautics and Aestronautics, Shanghai Jiao Tong University}
  \city{Shanghai}
  \country{China}
  \postcode{200240}
}



\begin{abstract}
Most current NeRF variants for satellites are designed for one specific scene and fall short of generalization to new geometry. Additionally, the RGB images require pan-sharpening as an independent preprocessing step.
This paper introduces psPRF, a Planar Neural Radiance Field designed for paired low-resolution RGB (LR-RGB) and high-resolution panchromatic (HR-PAN) images from satellite sensors with Rational Polynomial Cameras (RPC).
%
To capture the cross-modal prior from both of the LR-RGB and HR-PAN images, for the Unet-shaped architecture, we adapt the encoder with explicit spectral-to-spatial convolution (SSConv) to enhance the multimodal representation ability.
To support the generalization ability of psRPF across scenes, we adopt projection loss to ensure strong geometry self-supervision.
The proposed method is evaluated with the multi-scene WorldView-3 LR-RGB and HR-PAN pairs, and achieves state-of-the-art performance. 
\end{abstract}



\keywords{Planar Neural Radiance Field, Pan-Sharpening, Multimodal Neural Radiance Field}

\begin{teaserfigure}
  \includegraphics[width=\textwidth]{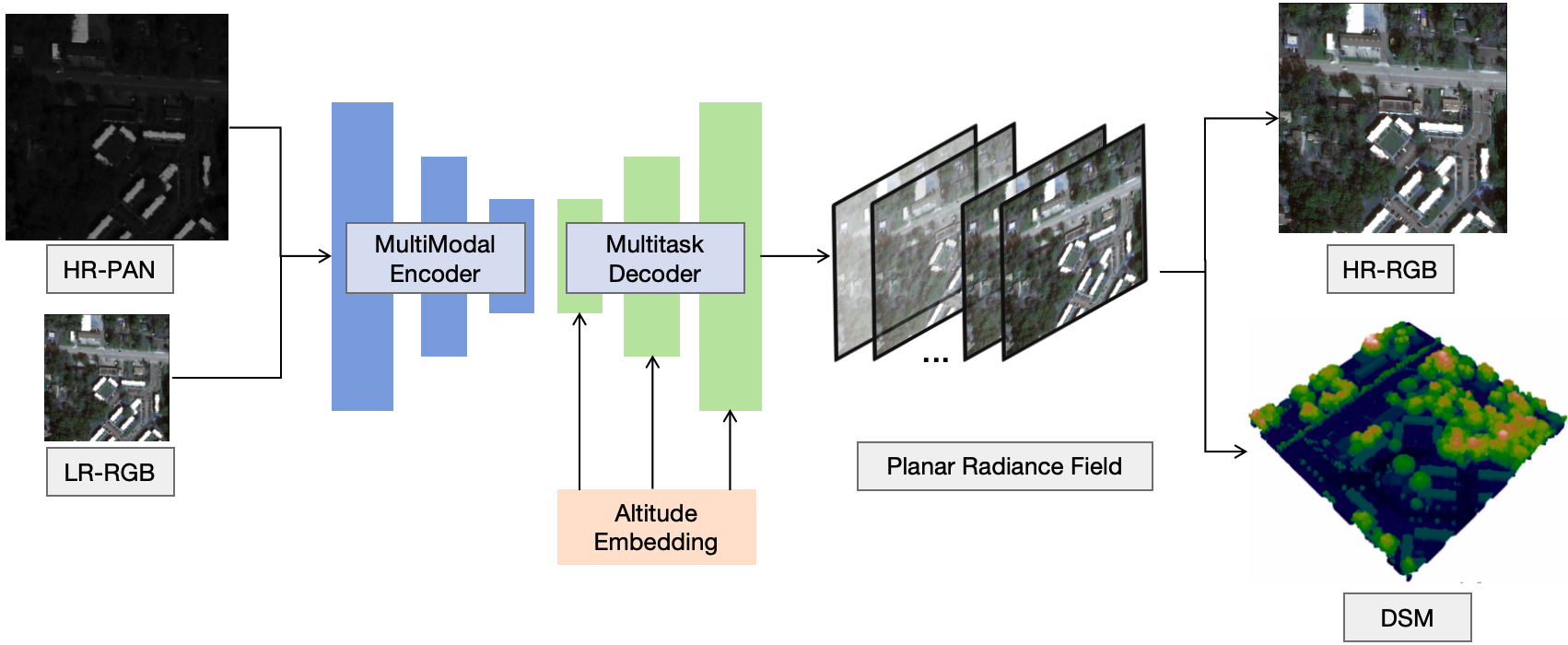}
  \caption{psPRF can produce high-resolution and novel views, as well as digital elevation models (DSMs) from a single pair of high-resolution panchromatic and low-resolution RGB images. It can also be utilized as a planar radiance field for sparse views in changing scenes.}
  \Description{}
  \label{fig:teaser}
\end{teaserfigure}


\maketitle

\section{Introduction}
\label{sec:intro}
Satellite sensors have a trade-off between spectral and spatial resolution. To compensate for the low-resolution optical RGB sensor, a single-band panchromatic (PAN) sensor with high spatial resolution can be referenced with finer details. The information from these sensors is usually combined offline through a process called pan-sharpening. This process creates the ideal full-resolution image for various tasks. Several optical Earth observation satellites, such as QuickBird, GeoEye-1, GaoFen-2, and WorldView-3, have been deployed to obtain panchromatic (PAN) and multispectral (MS) imagery containing RGB bands simultaneously.
Pan-sharpening uses a higher-resolution panchromatic image (or raster band) to fuse with a fully overlapped lower-resolution multiband raster dataset. The result produces a multiband raster dataset with the resolution of the panchromatic raster \cite{ps_review_jstar,ps_benchmark,ps_review_isprs_2021}.
%
High-resolution PAN (HR-PAN) and low-resolution RGB (LR-RGB) are often provided together in pairs. Most pan-sharpening methods treat these two tasks as independent computer vision problems without considering 3D geometry.
%
%
However, if additional camera information is available, it is promising to jointly reconstruct the 3D scene when accomplishing the pan-sharpening task.

For space-born 3D earth observation, advancements in deep learning have brought about a revolution in Multi-View Stereo (MVS). This is achieved by depth or normal regression \cite{mvsnet} with differentiable cost volumes or patchmatch step \cite{pmnet}.
In the meantime, implicit neural representations (INR) have made significant advancements in 3D tasks, particularly with the development of Neural Radiance Fields (NeRF) \cite{nerfreview}. This has led to a new revolution in unsupervised 3D reconstruction.
For the novel view synthesis (NVS) task, NeRF implicitly fits the color for every sampled volume and synthesizes the pixel color for the image under any novel viewpoint from limited images within one scene.
For the 3D reconstruction task, NeRF streamlines the traditional pipeline of matching tie points and triangulation without dense depth supervision \cite{chen2021mvsnerf,nerfingmvs}.

%
With the advancement of NeRF, more variants have been developed for various tasks, especially for satellite images. Sat-NeRF \cite{satnerf} and EO-NeRF are designed for cropped multi-date images for one scene with Rational Polynomial Cameras (RPC).  SatensoRF \cite{satensorf} adapts for multi-date images with original large sizes, but still for single scene.
The architecture of the vanilla NeRF is limited to a single scene with sufficient images.
To enhance the generalization ability 
The deep latent prior can be constructed by cost volume \cite{chen2021mvsnerf,nerfingmvs}, aggregated image features \cite{grf,ibrnet}, and some works adopt cross attention across views \cite{gpnr,gnt,gnt-move}.
Currently, all the GeNeRFs are designed for pin-hole optical sensors, for both monocular and sparse-view image settings.

For satellite images with RPCs, the rpcPRF model \cite{rpcprf} adopts the planar neural radiance field and projection to achieve self-supervision in altitude estimation. However, the input image resolution still affects the final results.
%
%
To date, most INR variants adopt the pan-sharpened RGB images as necessary high-resolution input, which treat pansharpening and color correction as independent pre-processing steps. 

The work aims to develop a generalized multimodal neural radiance field, named psPRF, for generalizable multimodal 3D reconstruction without the need for pan-sharpening in advance. 
 The multimodal encoder aligns spectral and spatial information from LR-RGB and HR-PAN pairs to model interactions across modalities, providing sufficient latent prior. 
As a radiance field, psPRF synthesizes high-resolution RGB imagery along with PAN imagery, treating pan-sharpening as a task of image synthesis. Therefore, based on the U-net-shaped MPI backbone, the decoder of the psPRF output three variables for every sampled pixel: the color intensity vector $\mathbf{c}_{RGB}$ of the RGB image, 
the intensity of the PAN image $c_{PAN}$, and the volume density $\sigma$.
psPRF samples the MPI frustum in the image space and apply differentiable RPC warping to project the object space to the image space, to enable differentiable MPI warping. 
The planar volume rendering for each frustum produces RGB color and altitude with the same resolution as the input HR-PAN. The design guarantees efficient inference and generalization ability, overcoming the limitations of EO-NeRF. 
The major contributions include:
\begin{itemize}
    \item We propose psPRF, a multimodal generalized planar neural radiance field for joint 3D reconstruction from single view HR-PAN and LR-RGB image pairs.
    \item Pan-sharpening has been accomplished as a byproduct of 3D reconstruction in the psPRF pipeline, which can be applied to other tasks.
    \item The experiments have validated the effectiveness of psPRF for sparse input from one scene, and monocular input from changing scenes.
\end{itemize}

\section{Related works}

\subsection{Pan-sharpening methods}
The traditional pan-sharpening methods are generally classified into three types. The first type, known as component substitution (CS) method, utilizes forward spectral transformation to replace specific spatial parts, ultimately generating the HR-MS image through an inverse spectral transformation.
The second type based on multi-resolution analysis (MRA) also substitutes components by wavelet transform and Laplacian pyramid, \etc \cite{ps_review_isprs_2021}, 
Variational optimization (VO)-based methods apply the variational theory with the deep image priors \cite{ps_review_jstar}.

With the development of computer vision methods, convolutional neural network (CNN) and the attention mechanism have been successively applied to pan-sharpening, such as PNN \cite{pnn}, SCPNN\cite{scpnn} and PanNet \cite{pannet}.
Generative models also inspire the pan-sharpening task such as PSGAN \cite{psgan}.
PSCINN \cite{ps_inn} applies a constraint-based invertible neural network between ground truth and LR MS images, conditioned by the guidance of the PAN image. 
UP-SAM \cite{ps_unsupervised} uses the self-attention mechanism to capture both spatial varying details and spectral characteristics, aiming to reduce the dependency on additional HRMS supervision. 
$\lambda$-PNN \cite{ps_sup_unsup} further explores the residual attention mechanism with novel loss design.
\subsection{NeRF for Satellite Images}
Vanilla NeRF \cite{nerfreview} is a system that uses MLPs to estimate the volume density and RGB color for each volume that is present in a complex scene. To capture the high-frequency components of the image signal, the scene is mapped to high-dimensional representations through positional encodings \cite{nerfreview}. Over recent years, several variants of the system have been customized for satellite applications.
SNeRF \cite{snerf} introduces the ambient hue composition model, while SatNeRF \cite{satnerf}  develops the sampling and rendering approach for a rational polynomial camera. SatensoRF \cite{satensorf} enhances light field composition and accelerates NeRF for satellite images using tensor decomposition. rpcPRF \cite{rpcprf} extends NeRF to different scenes by utilizing deep image features to create MPIs and incorporating geometry projection for guidance.
EO-NeRF \cite{eonerf} flexibly adjust rays to avoid bundle adjusting in advance, but still require pan-sharpened images as input.
\subsection{Prior-based Neural Radiance Field}
Incorporating deep image prior largely improves the generalization ability of NeRF models. SRF \cite{srf} and RUST \cite{sajjadi2023rust} use a vision transformer to infer a set-latent scene representation and then parameterize the light fields with the representation.
%
MVSNeRF \cite{chen2021mvsnerf} and its following works \cite{nerfingmvs,johari2022geonerf} are based on explicit cost volumes for post-regularization. The MVS-based methods apply to sparse input views well but require sufficient overlapping between the source and the reference views. 
More recently, more and more works have introduced transformers into NeRF pipelines.
IBRNet \cite{ibrnet} aggregates information with a transformer on the rays and conditions the Nerf decoder with the learned features. GNT \cite{gnt} and GPNR \cite{gpnr} implement the two-stage Nerf pipeline with a pure transformer-based model, with features accumulated along the epipolar lines and aggregated to estimate the color according to the view directions. 
 MatchNeRF \cite{matchnerf} uses an optical flow transformer to enhance cross-view interaction among views for correspondence matching.

\section{Preliminary}
\begin{figure*}
    \includegraphics[width=16cm]{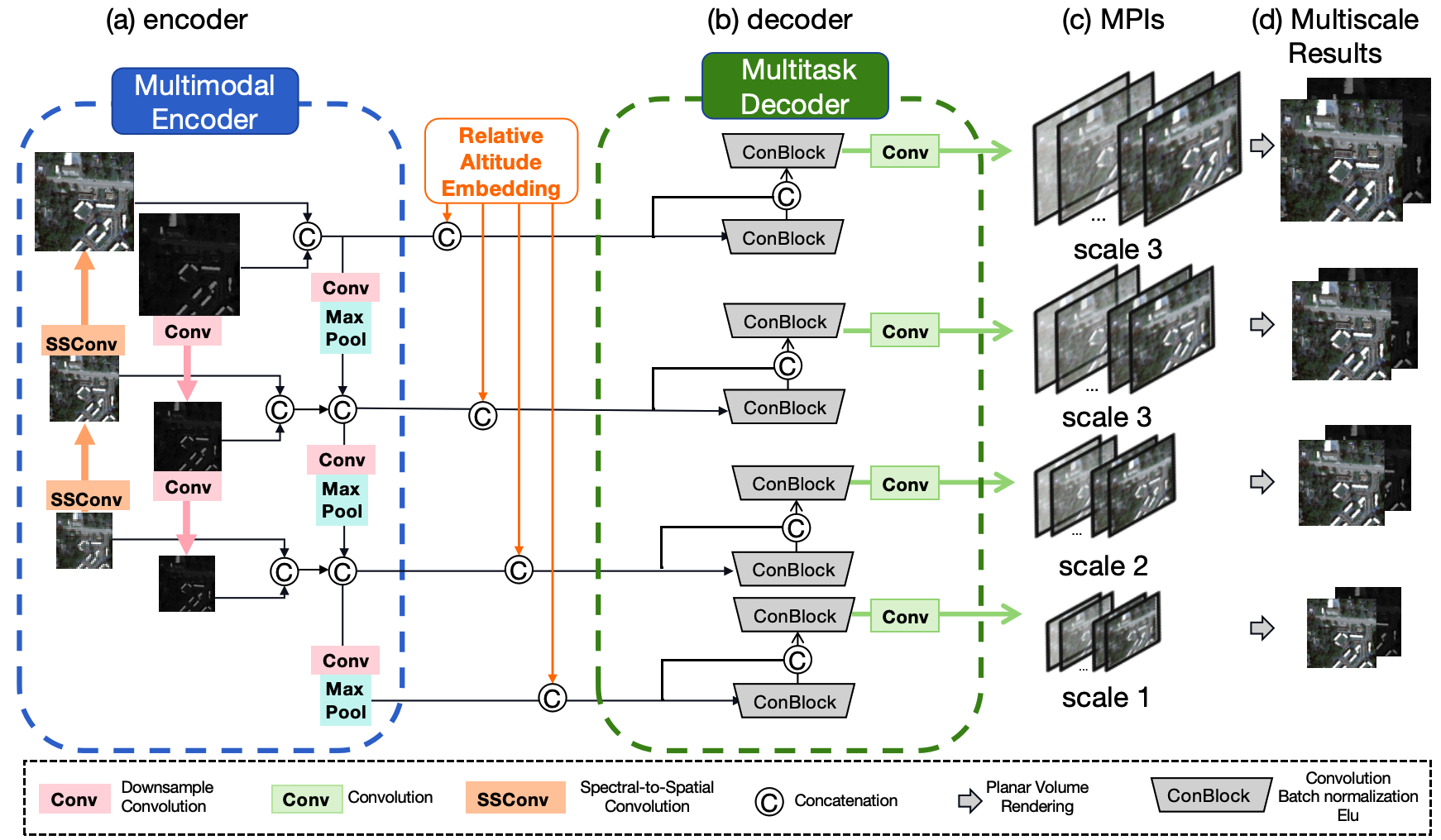}
    \caption{The pipeline of psPRF with a multimodal encoder, a multitask encoder to produce multiscale MPIs.}
    \label{fig:pipeline}
  \Description{}    
\end{figure*}
\subsubsection{Multiplane Image and Planar Volume Rendering}
Multiplane Image (MPI) uses a set of fronto-parallel planes at a fixed range of depths to generate a global scene.
The theoretical analysis by \cite {pushMPI} suggests that higher sampling frequencies can better represent stereo, and vanilla MPI can be extended to continuous 3D space at arbitrary depths by introducing neural rendering techniques. 
The planar neural radiance field adapts the volume rendering techniques for MPIs by assuming the sampling and simplified rendering integration in the continuous depth space. 
Given the RGBA set $\{(C_{h_i}, \sigma_{h_i})\}$ at the sampled height ${h_i}, i\in \{1, 2, \ldots, N_s\}$ for pixel $(x, y)$, $C_{h_i}$ being the color and $\sigma_{h_i}$ being the volume density, the rendering $\mathcal{PR}$ of a planar radiance field can be written as:
\begin{equation}
 \begin{aligned}
    \left\{
    \begin{array}{cc}
   & \hat{I}(x, y) = \sum\limits_{i=1}^NT_i(x, y)(1-\exp(-\sigma_{h_i}(x, y)\delta_{h_i}))C_{z_i}(x, y)\\
      &  \hat{H}(x, y) = \sum\limits_{i=1}^NT_i(x, y)(1-\exp(-\sigma_{h_i}(x, y)\delta_{h_i}))h_i
\end{array}
    \right. \label{eq:volrender}
\end{aligned}   
\end{equation}
where $\delta_{h_i}(x,y) = \|\mathcal{P}(x, y, h_{i+1})^\top-\mathcal{P}(x, y, h_i)^\top\|_2$
denotes the distance between the $i^{th}$ plane at height $h_i$ to the $(i+1)^{th}$ plane at height $h_{i+1}$ in the camera coordinate, with  $\mathcal{P}(\cdot)$ being the conversion from perspective 3D coordinate to the Cartesian coordinate. 
$T_i(x,y) = \exp(-\sum\limits_{j=1}^{i-1}\sigma_{h_i}(x,y)\delta_{h_j}), x \in [0, W], y \in [0, H]$ of an image with size $H\times W$ 
 indicates the accumulated transmittance from the first plane $\mathcal{P}(x, y, h_1)$ to the $i^{th}$ plane $\mathcal{P}(x, y, h_i)$, \ie,  the probability of a ray travels from $(x, y, h_1)$ to $(x, y, h_i)$ without hitting the object on the ground surface. 

\subsection{Rational Polynomial Camera Warping}
\label{sec:rpc_warp}
In the pipeline of MINE and its following variants, the frustum sampling is followed by homography warpings to the target novel views, with the given camera information.
For satellites with RPCs, rpcPRF \cite{rpcprf} proposes to differentiably warp the frustum with RPC coefficient tensor contractions to the target views in the Geodetic Coordinate System (GCS).

The RPC model 
is commonly used in high-resolution satellite photogrammetry due to its high approximation accuracy and the independence of the sensor and platform. 
%
It approximates the mapping between a 3D point in the normalized world coordinate and its projected 2D point on the sensor's image using ratios of polynomials represented by tensors:
\begin{equation}
    \left\{
    \begin{aligned}
    samp &= \frac{P_1(lat, lon, hei)}{P_2(lat, lon, hei)}=\frac{\mathcal{A}_{(num)}\mathbf{X}}{\mathcal{A}_{(den)}\mathbf{X}}\\
    line &= \frac{P_3(lat, lon, hei)}{P_4(lat, lon, hei)}=\frac{\mathcal{B}_{(num)}\mathbf{X}}{\mathcal{B}_{(den)}\mathbf{X}}\label{eq:rpc_forward}
    \end{aligned}
    \right. 
\end{equation}
where $lat$, $lon$, $hei$ denote the normalized latitude, longitude and height of the 3D point, $samp$ and $line$ are the normalized row and the column of the pixels, and for this point in the object space, its cubic tensor $\mathbf{X} = \{\mathbf{X}_{ijk}\}$ has rank 3, where $\mathbf{X}_{ijk} = [1, hei^i, lat^j, lon^k], i,j,k\in \{0, 1, 2, 3\}$.
And $P_t(\cdot), t\in \{1, 2, 3, 4\}$ represents the cubic polynomials to fitting the projection process from the world to the image. 
For instance, the polynomial mapping $P_1(\cdot)$ is formalized as:
\begin{align}
    P_1(lat, lon, hei) &= \mathcal{A}_{(num)}\mathbf{X} \nonumber\\ &= \sum\limits_{i=0}^{3}\sum\limits_{j=0}^{3}\sum\limits_{k=0}^{3}\mathcal{A}^{ijk}_{(num)}hei^ilat^jlon^k,\label{eq:rpc_coef}
\end{align}
The batched version of the projection in  Eq.(~\ref{eq:rpc_forward}) denoted as $\mathcal{F}_{proj}$, can be utilized to project the ground points tensor $\mathcal{G}$ from the object space, with the batched coefficient tensor $\mathcal{A}_{(num)}$, $\mathcal{A}_{(den)}$, $\mathcal{B}_{(num)}$ and $\mathcal{B}_{(den)}$:
\begin{align}
    \mathbf{samp}  = \mathcal{A}_{(num)}\mathcal{G} \oslash  \mathcal{A}_{(den)}\mathcal{G}\\ \notag
    \mathbf{line} = \mathcal{B}_{(num)}\mathcal{G} \oslash  \mathcal{B}_{(den)}\mathcal{G}\\ \notag
    \mathbf{samp}, \mathbf{line} \triangleq \mathcal{F}_{proj}(\mathcal{G},RPC)
  \label{eq:ten_proj}
\end{align}
where the batched ground point tensor $\mathcal{G}^{bn}_{ijk} = \{1, (hei^{bn})^i$, $(lat^{(bn)})^j$, $(lon^{(bn)})^k\}$ denotes the 3D points in the object space, with $i, j, k\in \{0,1,2,3\}$ indexing the exponents, and $\oslash$ denotes the element-wise division.
For localization, \ie mapping from image to the ground, the inverse process $P_t(\cdot), t \in \{5, 6, 7, 8\}$ can also be approximated by polynomial ratios represented by tensor contraction. The inverse coefficient tensors as the numerator and denominator are denoted as $\mathcal{C}_{(num)}$, $\mathcal{C}_{(den)}$, $\mathcal{D}_{(num)}$, $\mathcal{D}_{(den)}\in \mathbb{R}^{4\times 4}$ respectively.  Then the cubic tensor $\mathbf{Y} = \{\mathbf{Y}_{ilm}\}$, with $\mathbf{Y}_{ilm} = [1, hei^i, samp^l, line^m], i,l,m\in \{0, 1, 2, 3\}$ denotes the point in the image space:
\begin{equation}
    \left\{
    \begin{aligned}
    lat &= \frac{P_5(samp, line, hei)}{P_6(samp, line, hei)}=\frac{\mathcal{C}_{(num)}\mathbf{Y}}{\mathcal{C}_{(den)}\mathbf{Y}}\\
    lon &= \frac{P_7(samp, line, hei)}{P_8(samp, line, hei)}=\frac{\mathcal{D}_{(num)}\mathbf{Y}}{\mathcal{D}_{(den)}\mathbf{Y}}
    \end{aligned}\label{eq:inv_rpc}
    \right. 
\end{equation}
The batched version of localization denoted as $\mathcal{F}_{loc}$, with batched RPC tensor $\mathcal{C}_{(num)}$, $\mathcal{C}_{(den)}$, $\mathcal{D}_{(num)}$,
$\mathcal{D}_{(den)}$ and batched MPI tensor $\mathcal{M}$ can be written as:
 \begin{align}
    \mathbf{lat} &= \mathcal{C}_{(num)}\mathcal{M} \oslash\mathcal{C}_{(den)}M \\ \notag
    \mathbf{lon} & = \mathcal{D}_{(num)}\mathcal{M}\oslash\mathcal{D}_{(den)}\mathcal{M}\\ \notag \mathbf{lat}, \mathbf{lon}&\triangleq\mathcal{F}_{loc}(\mathcal{M},RPC) \label{eq:ten_loc}
\end{align}
where $\mathcal{M}^{bn}_{ijk} = \{1, N^{(bn)i}, samp^{(bn)j}, line^{(bn)k}\}$ denotes the pixels of the MPI tensor $\mathcal{M}$ in the image space,  with $i, j, k\in \{0,1,2,3\}$ indexing the exponents.

\section{Proposed Method}
\subsection{Problem Formulation and Pipeline}
We first formulate the problem and introduce the architecture, with module details explained in the following sections. 
Given a LR-RGB $I_{LR-RGB} \in \mathbb{R}^{w\times h\times 3}$ and its corresponding HR-PAN $I_{HR-PAN} \in \mathbb{R}^{W\times H\times 1}$ with the augmented RPC tensor $RPC$, the goal of the network is to predict $N$ planes:
\begin{equation}
    \{(c_i, p_i, \sigma_i)\}_{i=1}^N = \mathcal{F}(I_{LR-RGB}, I_{HR-PAN})
\end{equation}
where $c_i$ is the RGB color of the $i_{th}$ plane to generate the HR-RGB $\hat{I}_{HR-RGB}$ via integration, $p_i$ is the panchromatic intensity to generate the HR-PAN $\hat{I}_{HR-PAN}$, and $\sigma_i$ is the plane density to generate the transmittance. The final product of psPRF is the synthetic HR-RGB $\hat{I}_{HR-RGB}$, the synthetic HR-PAN $\hat{I}_{HR-PAN}$ and the DSM converted from the estimated altitude estimation $\hat{H}$. The integration is approximated by Eq.(~\ref{eq:volrender}).

To encompass multiscale levels of details for the pixel-wise task,  the model $\mathcal{F}$ is implemented in the U-Net shape with a multimodal encoder $\mathcal{F}_{enc}$ and a monodepth2 decoder $\mathcal{F}_{dec}$, shown in Fig.~\ref{fig:pipeline} (a).

\subsection{Multimodal Encoder}
To aggregate the spatial information of HR-PAN and the spectral information of LR-RGB, 
the features of the two input modalities should be concatenated with different scales though the encoder.
To align the two modalities in the feature space via concatenation, LR-RGB should be upsampled and HR-PAN should be downsampled respectively.
Specifically, to update LR-RGB, we adopt the Spectral-to-Spatial Convolution (SSConv) \cite{ssconv} as the upsampling operator to fuse the spectral and spatial information. 
For the feature map $\mathbf{F}\in\mathbb{R}^{w\times h\times c}$, the output feature map $O_i = \mathbf{F}\otimes K_i, i\in \{0, 1, \ldots, r^2c-1\}$ is defined as 
\begin{align}
    &O_{i, j, kr^2 + c_1r + c_2} = SSconv(\mathbf{F})_{ri+c_1, rj + c_2, k}\\
    &i\in \{0, \ldots, w-1\}, j\in \{0, \ldots, h-1\},\\
    &c_1\in \{0, \ldots, r-1\}, c_2\in \{0, \ldots, r-1\},\\
    & k\in \{0, \ldots, c-1\}
\end{align}
where $K_i\in \mathbf{R}^{1\times 3\times 3\times c}$ is the $i_{th}$ convolution kernel,  $r=2$ is the upsample ratio.

\begin{figure*}
    \centering
    \includegraphics[width=15cm]{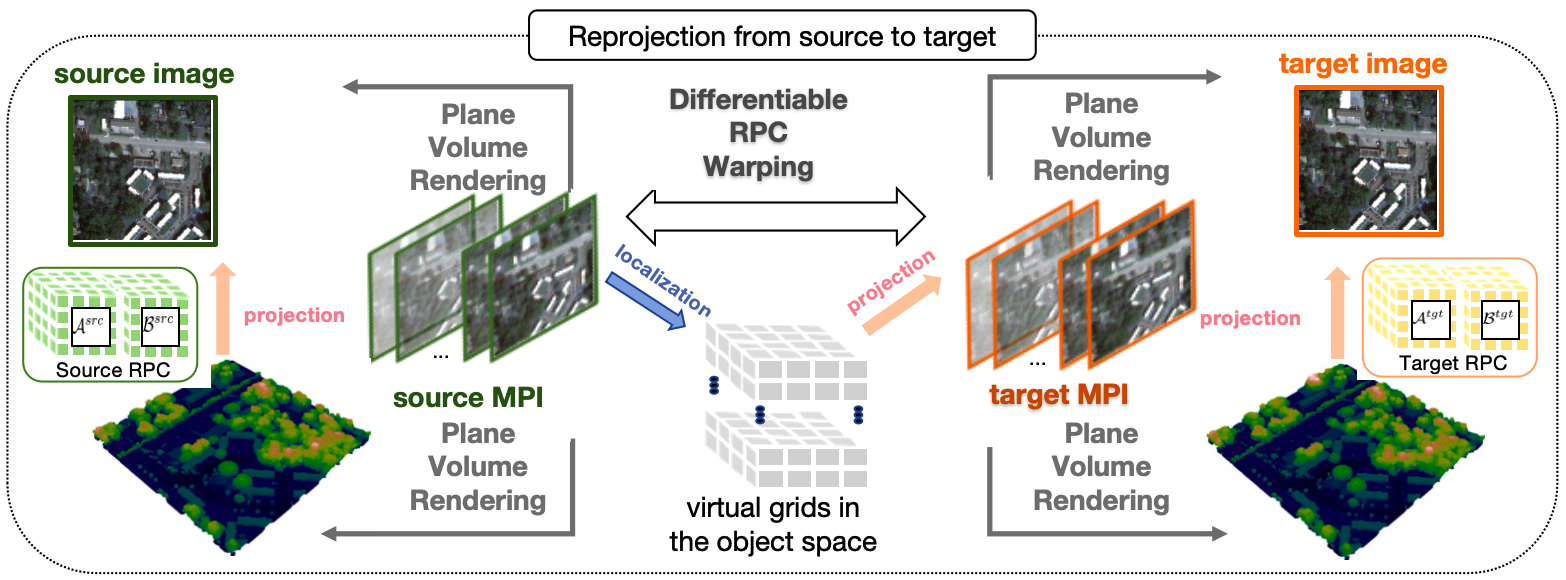}
    \caption{Reprojection from the synthesized product of the source view to the products of the target view by warping the frustum.}
    \label{fig:src2tgt}
     \Description{}  
\end{figure*}

\subsection{MPI Decoder with Depth Embedding}
To achieve accurate sampling in the object space, it is desired to sufficiently utilize the sampled height hypotheses between the lowest altitude $h_{far}$ and the highest altitude $h_{near}$ in the altitude space. 
Different from common MPIs and planar neural radiance fields, the height range $[h_{near}, h_{far}]$ is much larger than common 3D scenes. Besides, the altitude range sometimes include the zero elevation, therefore we take the uniform altitude sampling rather than in the inverse altitude space.

Additionally, to fully represent the comprehensive features of the large-scale height space when initializing the MPI space, we utilize a 1-dimensional positional embedding for the relative height sample index $ h_i, i \in \{ 0, 1, 2, \ldots , N_s-1 \} $ by frequency encoding:
\begin{align}
       \gamma(h_i) = & [\sin(2^0\pi h_i), \cos(2^0\pi h_i),\ldots,\nonumber \\ 
   & \sin(2^{L-1}\pi h_i), \cos(2^{L-1}\pi h_i)]\label{eq:posit_enc} 
\end{align}
The altitude embeddings are concatenated with multiscale features extracted by the encoder at a later stage, as part of the features in the altitude space, as shown in Fig.~\ref{fig:pipeline} (b)

\subsection{Reprojection via RPC Warping}

\begin{figure}[t]
    \centering
    \includegraphics[width=8cm]{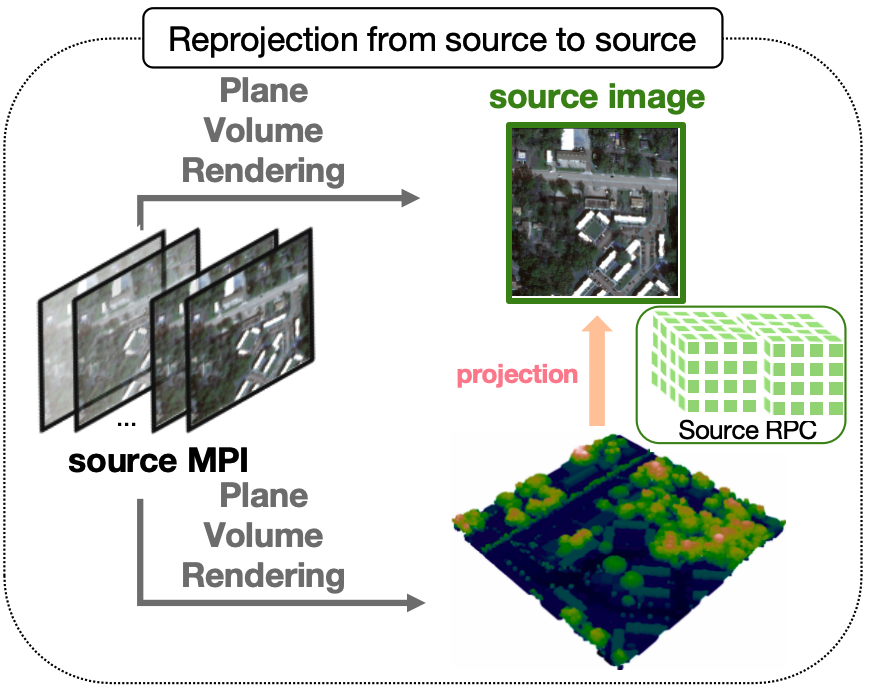}
    \caption{Reprojection from the synthesized altitude map to the rendered image of the source view.}
    \label{fig:src2src}
       \Description{}  
    \vspace{-0.8cm}
\end{figure}

To explore the geometry between the image space and the object space, psPRF adopts reprojection loss as self-supervision.
As shown in Fig.~\ref{fig:src2src}, in the forward process of the pipeline, with the corresponding projecting RPC tensor $RPC^{src}$, the projection of the synthesized altitude map $\hat{H}$ yields another projected image $I_{proj-RGB}$ manifesting the rigorous image and object correspondence.
 and then $\mathcal{G}$ is projected back to the source MPI with the source RPC:
  \begin{align}
     samps, lines &= \mathcal{F}_{loc} (\hat{H}, RPC^{src})\\ \notag
     \mathcal{M}^{src}_{proj} &= [samps^{src}, lines^{src}, \hat{H}]\\ \notag
     I_{proj-RGB}^{src} &= \mathcal{PR}(\mathcal{M}^{src}_{proj})
     \label{eq:loc_grid_src}
 \end{align}
where $\mathcal{PR}$ indicates the planar volume rendering.

To encourage the model to fit the correct geometry, the single-view reprojecting loss can be given by summing up the multiscale losses between the projected HR-RGB and the source HR-RGB:
\begin{equation}
    \mathcal{L}_{src-reproject} = \sum\limits_{s=1}^4\sum\limits_{n=1}^N\|I_{proj-RGB}^{src}-I_{HR-RGB}^{src}\|_1
\end{equation}.

In the sparse view settings, the model is fed with a group of images containing one HR-PAN $I_{HR-PAN}^{src}$, one $I_{LR-RGB}^{src}$, and $N_{view}-1$ target views: $\{I_{LR-RGB}^{tgt(1)}, \ldots, \{I_{LR-RGB}^{tgt(N_{view}-1)}\}$.
 As illustrated in Fig.~\ref{fig:src2tgt}, the reprojection from source image to the target image requires differentiable warping of MPIs.
 MPI warping includes two steps, firstly the source MPIs are mapped to the virtual grids $\mathcal{G}$ in the object space via  localization with the augmented part of the source RPC by applying $\mathcal{F}_{proj}$ to the source MPI $\mathcal{M}^{src}$ as:
 \begin{align}
     lats, lons &= \mathcal{F}_{loc} (\mathcal{M}^{src}, RPC^{src})\\ \notag
     \mathcal{G} &= [lons,lats, \hat{H}]  
     \label{eq:loc_mpi}
 \end{align}

 and then $\mathcal{G}$ is projected to the target MPI with the target RPC:
  \begin{align}
     samps, lines &= \mathcal{F}_{loc} (\mathcal{G}, RPC^{tgt})\\ \notag
     \mathcal{M}^{tgt} &= [samps^{tgt}, lines^{tgt}, \hat{H}] \\ \notag
        I_{proj-RGB}^{tgt} &= \mathcal{PR}(\mathcal{M}^{tgt}_{proj})
     \label{eq:proj_grid_tgt}
 \end{align}

To guarantee the epipolar geometry, the sparse-view reprojecting loss can be given by summing up the multiscale losses:
\begin{equation}
    \mathcal{L}_{tgt-reproject} = \sum\limits_{s=1}^4\sum\limits_{n=1}^N\|I_{proj}^{tgt}-I_{HR-RGB}^{tgt}\|_1
\end{equation}.

\subsection{Training Loss}
To ensure the synthesized HR-PAN image's photometric fidelity with the ground truth HR-PAN, we adopt a panchromatic fidelity term:
\begin{equation}
    \mathcal{L}_{pan} = \sum\limits_{s=1}^4\sum\limits_{n=1}^N\|\hat{I}_{pan}-\mathcal{I}_{HR-PAN}\|_1
\end{equation}
Similarly, to encourage the photometric fidelity of the synthesized RGB image and the ground truth high-resolution RGB image (HR-RGB), L1 loss can be adopted when the HR-RGB is provided:
\begin{equation}
        \mathcal{L}_{color} = \sum\limits_{s=1}^4\sum\limits_{n=1}^N\|\hat{I}_{RGB}-I_{HR-RGB}\|_1
\end{equation}
Therefore, the total training loss is composed of:
\begin{equation}
    \mathcal{L}_{total} = \lambda_1 \mathcal{L}_{pan} +  \lambda_2 \mathcal{L}_{color} +  \lambda_3 \mathcal{L}_{reproject}
\end{equation}
where $\lambda_1, \lambda_2, \lambda_3$ indicate the ratios of the three losses.
$\mathcal{L}_{reproject} = \mathcal{L}_{src-reproject}$ for single view psPRF, and $\mathcal{L}_{reproject} = \mathcal{L}_{tgt-reproject}$ for sparse view psPRF.
\section{Experiments}
\begin{figure*}
    \includegraphics[width=15cm]{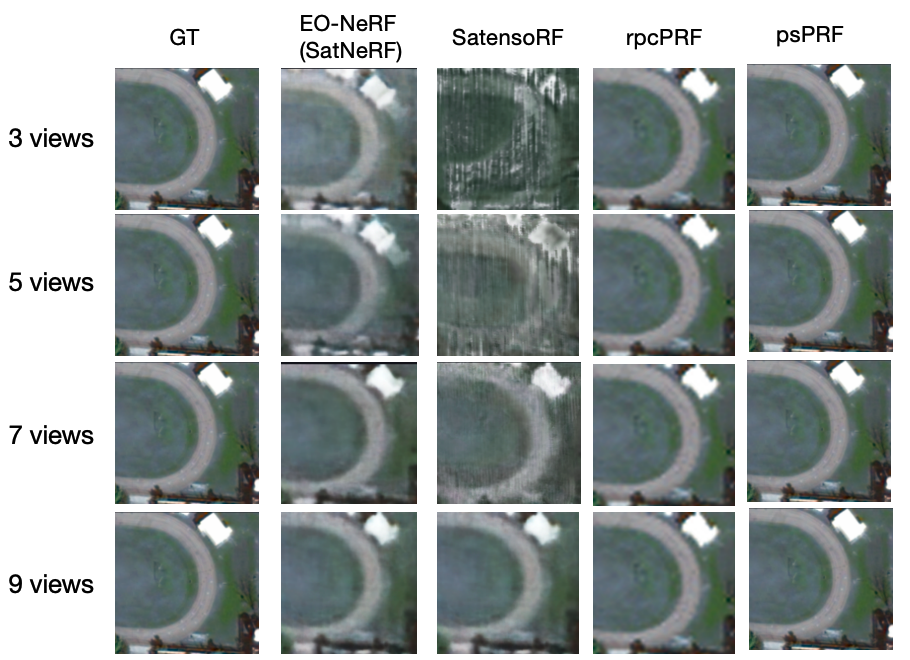}   
    \caption{When all the input images are from a single scene, psPRF is able to produce satisfactory results even when there are very few input views. However, the performance of EO-NeRF (SatNeRF) and SatensoRF are severely affected when the number of input views is reduced. While rpcPRF is robust to view reduction, it fails to enhance the resolution according to input HR-PANs.}
    \label{fig:comp_one_scene}
       \Description{}  
\end{figure*}
\begin{figure*}
\centering
    \includegraphics[width=15cm]{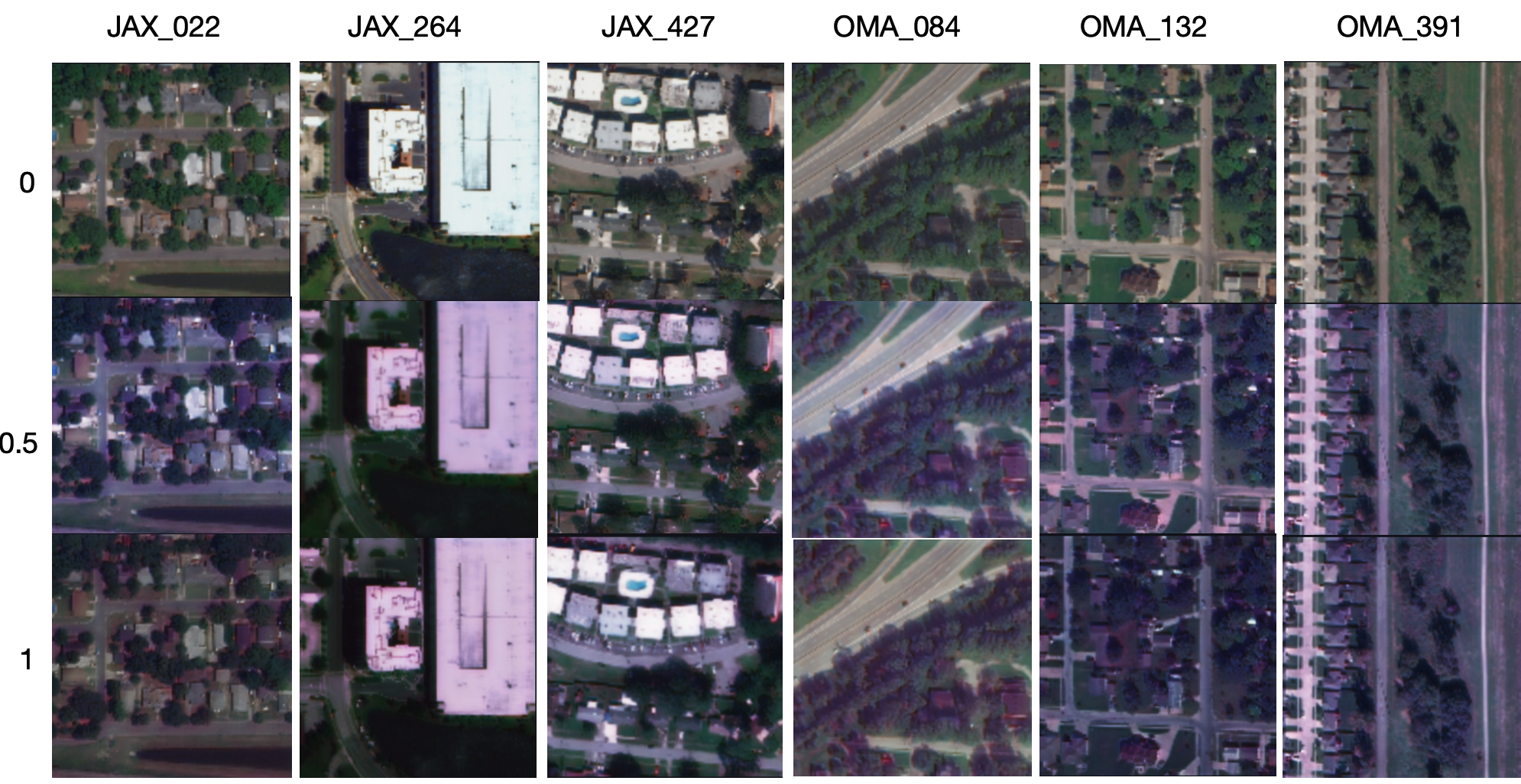}
    \caption{Ablation study of how different level of reprojection loss affects the rendering results.}
    \label{fig:ablation_depth}
     \Description{}  
\end{figure*}
\subsection{Experimental Setup}
\subsubsection{Evaluation Goals}
To validate the generalization ability of psPRF compared to the State-Of-The-Art (SOTA) satellite 3D reconstruction models, we adopt the evaluation settings of both the canonical NeRF and the monocular reconstruction, the evaluation protocol and the roadmap are as follows:
\begin{itemize} 
    \item \textbf{NeRF for one scene}: The training, test, and validation set are selected from the same scene, with all the input images coming in groups of multi-view RGB images and their corresponding RPCs.
    For NeRF in single scene, the experiment is given in Section~\ref{sec:comp_nerf_one_scene}.
    For fairness, the RGB images are not downsampled for all the models.
\item \textbf{Monocular reconstruction for changing scenes}: The training, test, and validation set are selected from different scenes, with all the input images coming in HR-PAN and LR-RGB pairs. For changing scenes, the experiments is given in Section~\ref{sec:comp_nerf_across_scene}.
To evaluate psPRF on processing the images without pan-sharpening, all the RGB images are downsampled four times to form the LR-RGB.
\end{itemize}

For novel view synthesis, we take the common image quality indicators, the learned perceptual image patch similarity (LPIPs), the peak signal-to-noise ratio (PSNR), and the structural similarity index measure (SSIM)  
 as metrics. 
\subsubsection{Data Source}
%
All the experiments are conducted on the images from the Maxar WorldView-3 satellite \footnote{https://spacenet.ai/iarpa-multi-view-stereo-3d-mapping/}. The images are collected from Jacksonville, Florida, and Omaha, Nebraska between 2014 and 2016.
Based on the processing of DFC2019 dataset (2019 IEEE GRSS Data Fusion Contest), we crop the original RGB and PAN images according to the region of interest (ROI) from the DSM file.
To enable the batched process of the convolution neural network, we crop the images from irregular shapes to $512\times 512$, and project the pixel boundaries back to GCS to obtain the corresponding cropped DSM.
%
      

\begin{table}
    \centering
        \caption{ To evaluate psPRF as a radiance field for one single scene,  four subsets from the Worldview3 dataset are chosen, with details listed below. 
        }
    \begin{tabular}{c|cccc}
    \toprule
   subset  & \#Images & \#train&\#test&bounds [$m$]  \\
    \midrule
      JAX\_033 & 8 &  7 &1 & [-26.1, 2.4] \\
      JAX\_070  &19 & 18&1& [-28.2, -1.9] \\
      OMA\_042   &39 &36&3& [286.8, 315.7] \\
      OMA\_181  &16 & 14&1& [270.0, 287.4] \\
      
    \bottomrule
    \end{tabular}

    \label{tab:data_one_scene}
  
\end{table}

\subsubsection{Implementation Details}
The model and the compared models are all trained on an NVIDIA RTX 3090 (24GB) GPU. For psPRF, we adopt Adam optimizer with a multistep scheduler for the learning rate. The initial learning rates for the encoder and decoder are set to be 0.0001 and 0.0002, and the milestones for the schedulers vary according to the number of parameters and the epoch settings. 

\subsection{Compare with Satellite Radiance Fields with Sparse Input in One Scene}
\label{sec:comp_nerf_one_scene}
\subsubsection{Details}
In this section, we compare the models using sparse training input views from a single scene, based on the vanilla settings. The comparison is conducted on four subsets of the Worldview3 images, with details in Table~\ref{tab:data_one_scene}.
To validate the robustness of psPRF against sparse input, in this experiment, training image groups are randomly selected from the four subsets
with view numbers to be $3, 5, 7, 9$. Specifically, rpcPRF and psPRF can be applied to monoview, and for subset JAX\_033 with only 8 images, we skip the sub-experiment with groups of 9.
To be specific, a group of images contains one HR-PAN and $N\in \{1, 3, 5, 7, 9\}$ LR-RGBs for training.
For the multiview cases, we adopt the multiscale reprojection losses to the target views $\mathcal{L}_{tgt-reproject}$ to establish the geometry among the views when $B\in\{3, 5, 7, 9\}$.
\begin{table*} 
      \caption{Novel view synthesis comparison with a varying number of views within one scene of the JAX\_070 subset.} 
  \fontsize{7}{7}
  \selectfont 
  \LARGE
  \centering
  \resizebox{16cm}{!}{
  \begin{threeparttable}  
    \begin{tabular}{c|cccc|cccc|cccc}  
    \toprule  
    &\multicolumn{4}{c}{\textbf{PSNR}$\uparrow$}&\multicolumn{4}{c}{\textbf{SSIM}$\uparrow$}&\multicolumn{4}{c}{\textbf{LPIPS}$\downarrow$}\cr  
    \cmidrule(lr){2-5} \cmidrule(lr){6-9}\cmidrule(lr){10-13}  
   views&EO-NeRF&SatensoRF&rpcPRF&psPRF&EO-NeRF&SatensoRF&rpcPRF&psPRF&EO-NeRF&SatensoRF&rpcPRF&psPRF\cr 
    \midrule  
        1 & - & -& 25.21&\textbf{26.18} &- & -& 0.85&0.85 & -& -& 0.31&\textbf{0.28} \\
    3 &  12.72 &12.78 & 26.27& \textbf{26.41}&0.57 & 0.14& \textbf{0.86}& 0.85&0.57 & 0.73& \textbf{0.28} & 0.29\\
    5 & 16.61& 12.82&27.03&\textbf{27.21}&0.61&0.18&0.87&\textbf{0.88}&0.49&0.71&\textbf{0.26}&0.27\\
    7 & 21.73& 14.13&27.09&\textbf{27.33}&\textbf{0.89}&0.35 & 0.87&0.88&0.26& 0.53&0.25&\textbf{0.24}\\
    9 & \textbf{27.65}&15.72& 27.59 & 27.62&\textbf{0.95}&0.84 & 0.89 & 0.88 & 0.23 & 0.36 & \textbf{0.22} & 0.25\\
    \bottomrule  
    \end{tabular}  
    \end{threeparttable}
    }
    \label{tab:comp_one_scene_nvs}
\end{table*}
\begin{table}
    \centering
        \caption{Source view synthesis comparison on HR-PAN, LR-RGB pairs over different scenes.}
    \begin{tabular}{c|ccc}
    \toprule
    & PSNR&SSIM&LPIPs\\
    \midrule
         rpcPRF& 22.96&0.75&0.31 \\
         psPRF&\textbf{26.83} &\textbf{0.83}&\textbf{0.19}\\
         \bottomrule
    \end{tabular}

    \label{tab:comp_across_scene_ps}
\end{table}
\begin{table}
    \centering
        \caption{Novel view synthesis comparison on HR-PAN, LR-RGB pairs over different scenes.}
    \begin{tabular}{c|ccc}
    \toprule
    & PSNR&SSIM&LPIPs\\
    \hline
         rpcPRF&22.15 &0.69& 0.34\\
         psPRF&\textbf{26.34} &\textbf{0.82}&\textbf{0.21}\\
         \bottomrule
    \end{tabular}

    \label{tab:comp_across_scene_nvs}
\end{table}

\begin{figure*}
    \centering
    \includegraphics[width=16cm]{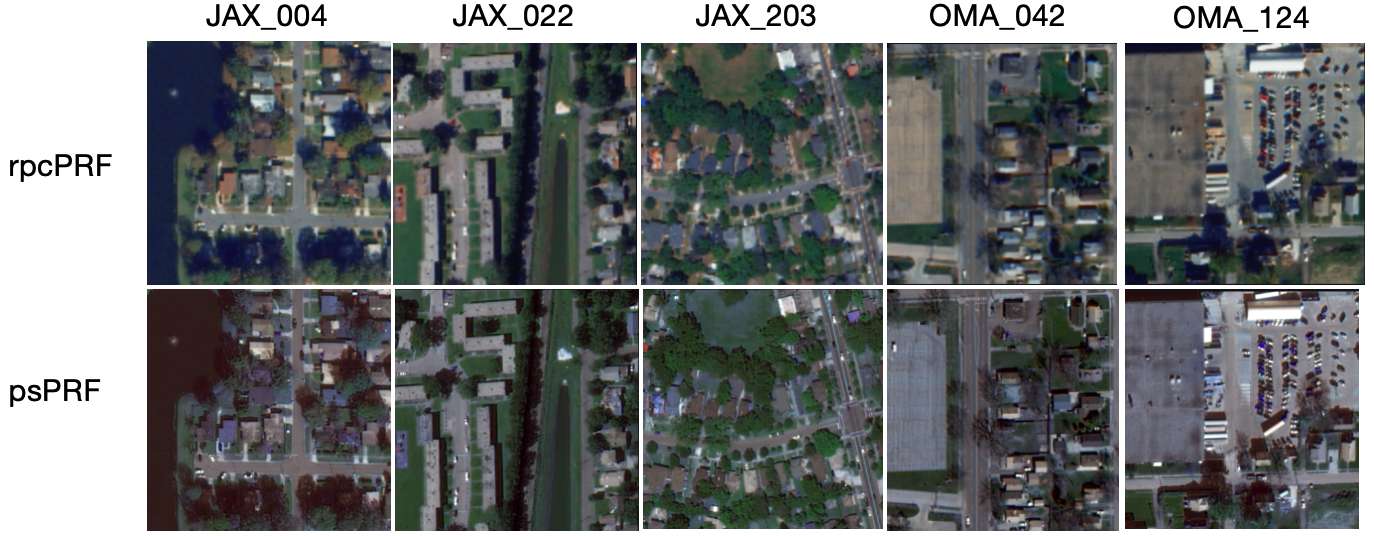}
    \caption{Comparison of the synthesizing results of the source image from single-view image pairs across different scenes.}
    \label{fig:comp_across_scene_ps}
       \Description{}  
\end{figure*}
\begin{figure*}
    \centering
    \includegraphics[width=15cm]{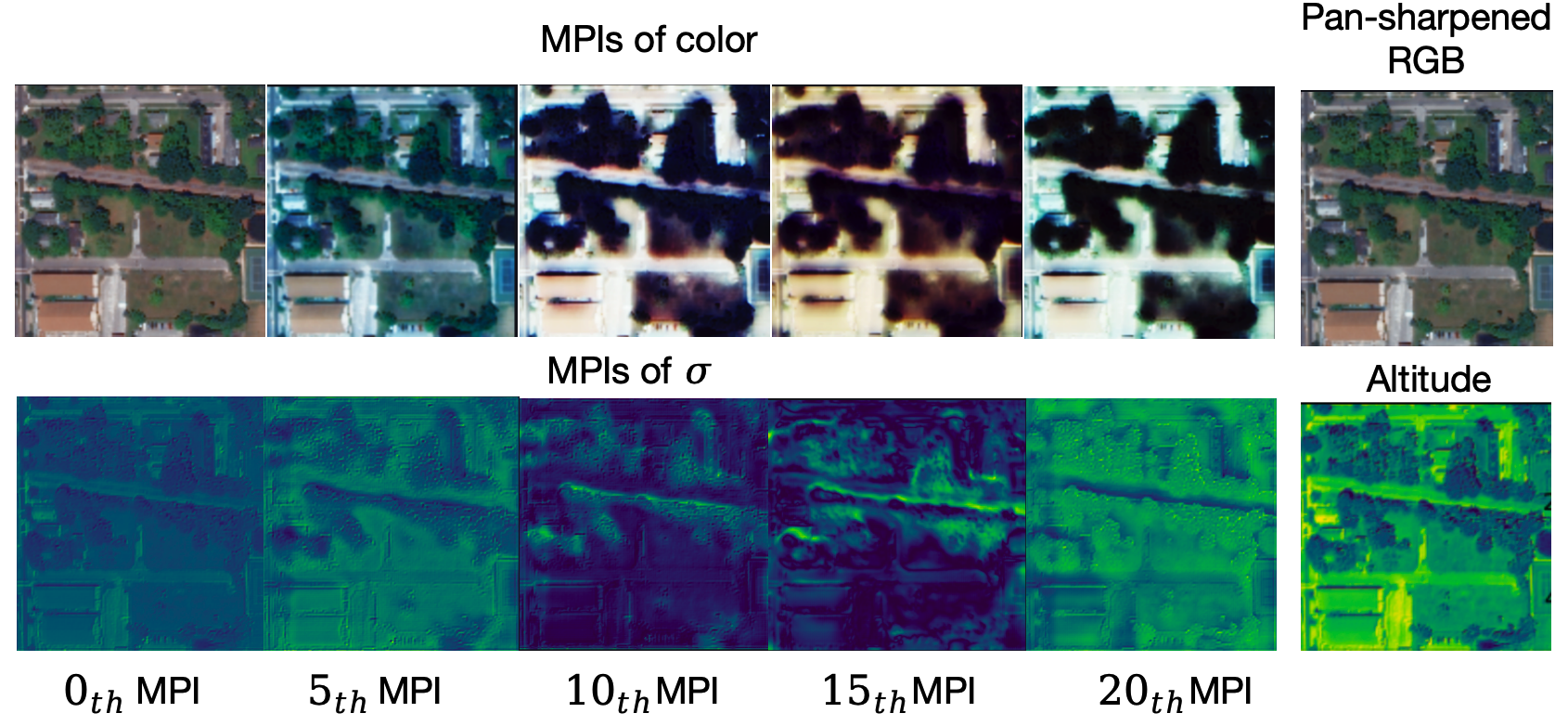}
    \caption{The MPI of the $15_{th}$ pairs of images of the JAX\_018 subset from the Worldview3 dataset.}
    \label{fig:mpis}
       \Description{}  
    \vspace{-0.5cm}
\end{figure*}

For comparison, the following models are selected as benchmarks:
\begin{itemize}
    \item EO-NeRF:     EO-NeRF is selected to be one comparison benchmark, as the most advanced derivation of Sat-NeRF.
    \item SatensoRF:  We select SatensoRF as a representation of the lightweight NeRF version for satellites, which requires no additional solar data.
    \item rpcPRF: As a planar neural radiance field for satellite, rpcPRF is selected for comparing the performance for sparse views as input.
\end{itemize}
 Due to the fact that psPRF requires no extra input of solar information, For fairness of input and computation, we removed the input of solar elevation and azimuth input for EO-NeRF but kept the embedding for transient objects.
\paragraph{results}  
  The quantitative comparison is recorded in Table \ref{tab:comp_one_scene_nvs}, where EO-NeRF (SatNeRF) slightly leads the performance when giving 9 views as input, however, the three metrics all deteriorate sharply when the number of view becomes less than 5.
  Meanwhile, rpcPRF and psPRF keeps the performance without big fluctuations. However, rpcPRF does not utilize the panchromatic images, thus yield lower resolution than results from the psPRF.
  The qualitative comparison is shown in Fig.\ref{fig:comp_one_scene}, where the upper-left corner of the $9_{th}$ image from subset JAX\_070 is selected and presented.

\subsection{Compare with Monocular Radiance Field Across Scenes}
\label{sec:comp_nerf_across_scene}
\paragraph{Details}
To validate the effectiveness of psPRF for single view generalization problem, we merged all the 105 subsets from the WorldView3 images together for the experiments. 
The training set and the test sets are randomly selected over every subsets, with 3980 pairs for training, 210 pairs for testing, and another 210 pairs for validation.
All the image pairs are compose of one HR-PAN, one LR-RGB, and the goal is to output the HR-RGB.

For comparison, only the monocular pipeline rpcPRF is selected as a benchmark.
Notably, rpcPRF here serves as a controlled experiment to validate the functionality of fusing spectral and spatial information.
There validations for pairs across different scenes are started from two perspectives:
\begin{itemize}
    \item  Evaluation of the pan-sharpening results: the input contains the pair of the source view  $I_{HR-PAN}^{src}, I_{LR-RGB}^{src}, RPC^{src}$, and output the pan-sharpened result $\hat{I}_{HR-RGB}^{src}$.    
    \item  Evaluation of the novel view synthesis results: the input contains the pair of the source view and the RPC of the target view $I_{HR-PAN}^{src}, I_{LR-RGB}^{src}, RPC^{src}, RPC^{tgt}$, and output the pan-sharpened result $\hat{I}_{HR-RGB}^{tgt}$.
\end{itemize}

\paragraph{result}
The qualitative comparison of pan-sharpening results is listed in Table~\ref{tab:comp_across_scene_ps}, the qualitative comparison of NVS results are listed in Table~\ref{tab:comp_across_scene_nvs},
and the comparison is shown in Fig.\ref{fig:comp_across_scene_ps}.
For source view reconstruction,\ie pansharpening the LR-RGB of the source view, psPRF yields finer details with better performances according to the metrics.

\subsection{Evaluation of the DSM}
After converting the height map to the DSM, we quantitatively evaluate the reconstruction quality using common metrics to assess altitude estimation:
\begin{itemize}
    \item The mean absolute error (\textbf{MAE}): The average of the L1 distance across all grid units between the ground truth and the estimated height map.
    \item Median Height Error (\textbf{ME}): The median of the absolute L1 distance over all the grid units between the ground truth and the estimated height map;
\end{itemize}
As an important byproduct, an accurate altitude map indicates the quality of NeRF model to learn the correct geometry.
\begin{figure*}
    \includegraphics[width=15cm]{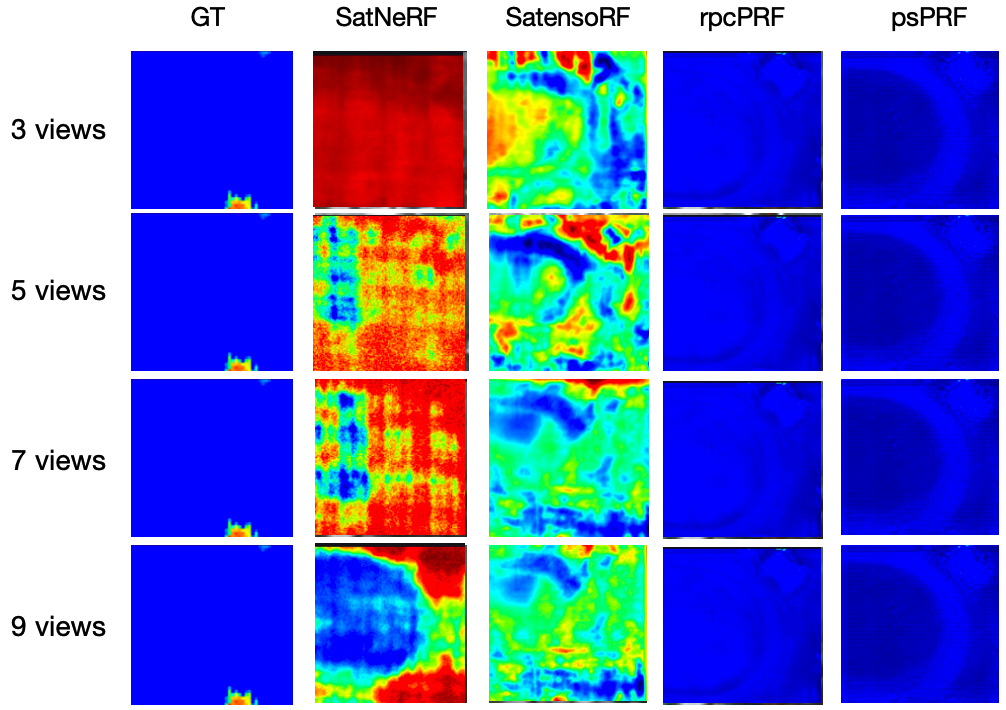}
    \caption{Altitude estimation comparison in the single scene JAX\_070 with a different number of input views.}
    \label{fig:comp_alt_single_scene}
     \Description{}  
\end{figure*}

We select the left-upper corner of scene JAX\_070 for visualization, following the image synthesizing paradigm.
In Fig.~\ref{fig:comp_alt_single_scene}, 
EO-NeRF achieves the best results in image synthesizing given 9 views but fails to estimate the actual altitude.
SatensoRF fails to learn the geometry given few views, while rpcPRF and psPRF learn the relatively correct geometry.
The quantitative comparison given 9 views is recorded in Table~\ref{tab:comp_alt_single_scene}. Due to the large altitude error calculated for EO-NeRF and SatensoRF, the converted DSM has a large bias from the original object coordinate with the less overlapped valid area. Therefore, the quantitative results of EO-NeRF and SatensoRF are left blank.
\begin{table}[H]
    \centering
     \caption{Unsupervised altitude estimation comparison of a single scene with different numbers of input views.}
     \resizebox{6cm}{!}{
    \begin{tabular}{c|cc}
    \toprule
     model&MAE(m)$\downarrow$& ME(m)$\downarrow$ \\
     \midrule
     EO-NeRF& - &-\\
     SatensoRF & - &-\\
     rpcPRF& 3.29 &3.15\\
     psPRF & \textbf{3.05} &\textbf{2.98}\\
    \bottomrule
    \end{tabular}   }
   
    \label{tab:comp_alt_single_scene}
\end{table}


\subsection{Efficiency}
\label{sec:exp_efficiency}
In addition to image quality evaluation,  we also compare the models by memory and computation consumption, time cost, and the model parameters at inference time. 
The three models are presented for comparison in Table~\ref{tab:comp_psprf_memory}.  

The two metrics \textbf{FLOPs} and \textbf{Time} are calculated for a triple of images. 
NeRF-based models are rendered per ray, so the temporary memory depends on the batch size of rays, while rpcPRF and psPRF are forwarded as image batches, so we set the ray batch size of EO-NeRF and SatensoRF to be $3 \times H \times W$ as a whole image.
%
\begin{table}[H]
    \centering
     \caption{Time and computation comparisons of benchmark methods and the proposed psPRF}
      \label{tab:comp_psprf_memory}
    \begin{tabular}{c|ccc}
    \toprule
        sites &  
        \textbf{Param}$\downarrow$& \textbf{FLOPs}$ (M)\downarrow$& \textbf{Time}$\downarrow$\\
        \midrule
EO-NeRF& 2301956&602167.8& 3.17 \\
SatensoRF&456780 &121131.5 &0.91 \\
rpcPRF&20299768& 202318.6 &0.08\\
psPRF&460002 &272785.4 &0.10 \\
         \bottomrule
    \end{tabular}

\end{table}

Table~\ref{tab:comp_psprf_memory} shows that the proposed psPRF enhances the inference speed by a large margin compared to EO-NeRF. 
Besides, the rendering process of the planar neural radiance field is performed per image rather than rendering per ray, which results in the efficient performance of psPRF.

\subsection{Ablation Study}
To investigate the influence of various factors for the proposed psPRF, we conducted the following experiments on the WorldView3 dataset with monocular reconstruction settings.

\subsubsection{Study on the reprojection loss}
\begin{figure*}
     \centering
     \includegraphics[width=15cm]{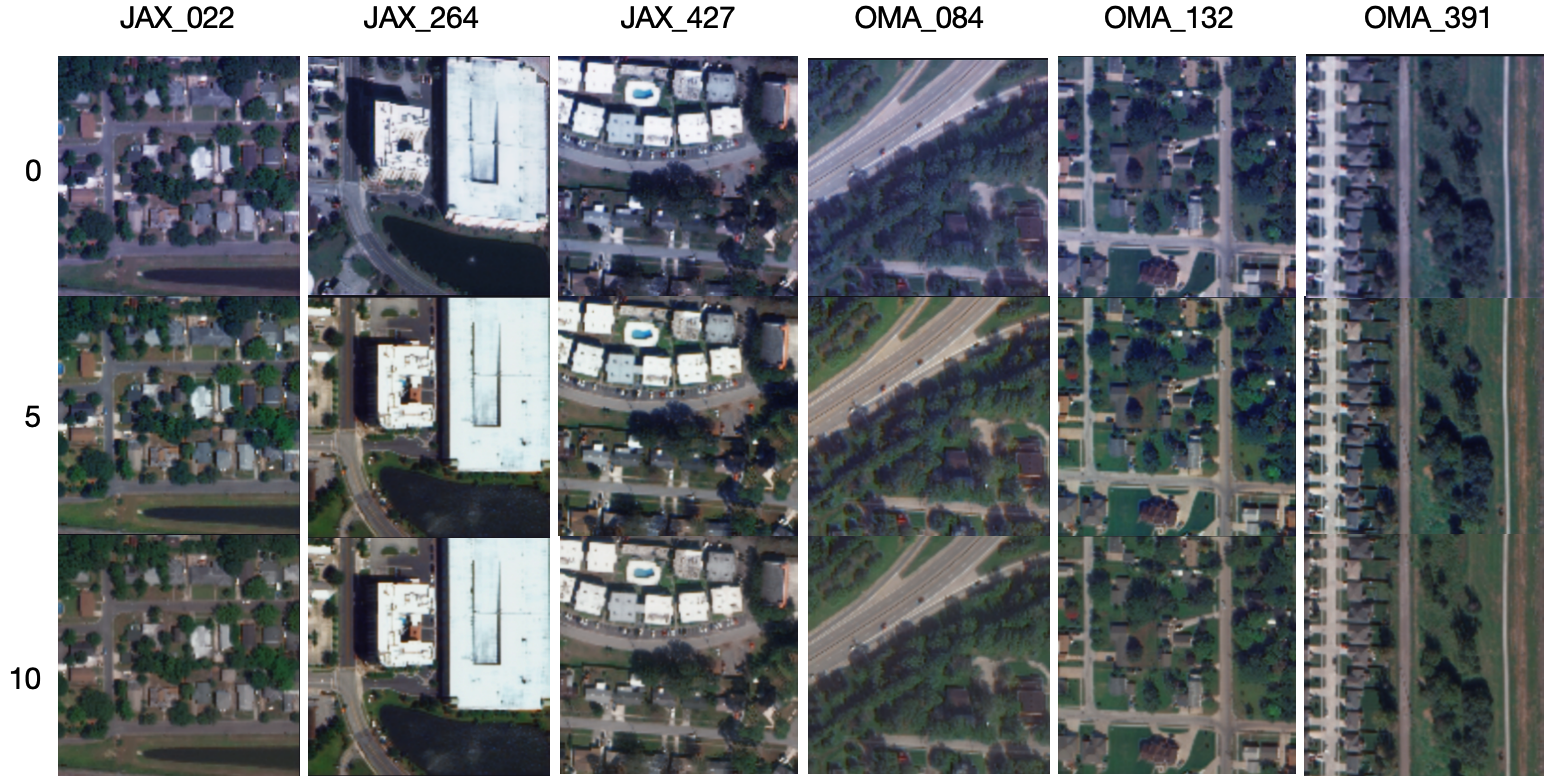}
     \caption{Ablation study of how different level of reprojection loss affects the rendering results.}
     \label{fig:ablation_reproj}
      \Description{}  
 \end{figure*}

This section inspects the availability of using the reprojection loss with different weights $\lambda_3\in \{0, 5, 10\}\}$ while $\lambda_1 = \lambda_2=1.0$.
When $\lambda_3 = 0$, there is no reprojection supervision exerted on psPRF.
\begin{table}[H]
    \centering
    \begin{tabular}{c|ccc}
    \toprule
        $\lambda_3$ & \textbf{PSNR}$\uparrow$ & \textbf{SSIM}$\uparrow$ & \textbf{LPIPs}$\downarrow$ \\
        \midrule
 0& 19.87&0.47 & 0.28  \\
  5& 24.60& 0.63& 0.30 \\
   10& 26.83 & 0.83& 0.19 \\
      
  \bottomrule
    \end{tabular}
    \caption{Investigation of how different level of reprojection supervision affects the results.}
    \label{tab:ablation_reprojection_psprf}
\end{table}
 the results are reported in Table~\ref{tab:ablation_reprojection_psprf}, and the qualitative comparison can be referred to Fig.~\ref{fig:ablation_reproj}, where the less accurate results without reprojection loss at the first row has color drifting away, which indicates the less accurate MPI and transmittance appears without geometric supervision.

\subsubsection{Study on the depth map supervision}
In the previous experiments, the reconstruction is conducted in an entirely unsupervised way.
This section inspects the utility of exerting depth map supervision on the psPRF model, and the reprojection loss is adopted when testing the depth loss.
\begin{table}
    \centering
    \begin{tabular}{c|ccc}
    \toprule
        $\lambda_3$ & \textbf{PSNR}$\uparrow$ & \textbf{SSIM}$\uparrow$ & \textbf{LPIPs}$\downarrow$ \\
        \midrule
  0&26.83 & 0.83&0.19  \\
  0.5&23.39 & 0.45& 0.35 \\
   1&22.11 & 0.38& 0.39 \\
  \bottomrule
    \end{tabular}
    \caption{Investigation of how different level of reprojection supervision affects the results.}
    \label{tab:ablation_depth_psprf}
\end{table}
Table~\ref{tab:ablation_depth_psprf} shows that applying the whole depth map as supervision severely affects the view synthesis results. 
Fig.~\ref{fig:ablation_depth} reveals the negative impact of depth loss on the image synthesizing performance.  

On the one hand, the discrepancy might be caused by the inaccuracy of the ground truth DSM for supervision.
On the other hand,
the negative impact might be caused by the unsynchronized optimization. For instance, the parameters converge for better transmittance with a faster speed than for better MPI weights.
Therefore, in future work, a more flexible MPI generator should be used to address the problem.

\section{Conclusion}

This paper proposes a joint pan-sharpening planar neural radiance field for sparse view or monocular satellite images, dubbed psPRF. As a generalization of MPI and neural rendering techniques, psPRF has better generalization ability over vanilla satellite NeRFs as a monocular model, which means it can be applied to pairs of high-resolution PAN and low-resolution RGB images pair from one view.
We explore the capability of the monocular model on the pans-sharpening problem. In the future, encouraging a more rigorous geometry model will be prioritized across different resolution by adapting the MPI decoder with a more unimodal distribution of the earth surface.





\bibliographystyle{ACM-Reference-Format}
\bibliography{sample-base}

\end{document}